\documentclass[runningheads]{llncs}
\usepackage{graphicx}
\usepackage{comment}
\usepackage{amsmath,amssymb} % define this before the line numbering.
\usepackage{color}

\usepackage[pagebackref=true,breaklinks=true,letterpaper=true,colorlinks,bookmarks=false]{hyperref}
\usepackage{epsfig}
\usepackage{enumitem}
\usepackage{multirow}
\usepackage[percent]{overpic}
\usepackage{xcolor}
\usepackage[margin=0pt]{caption}
\usepackage[margin=0pt,labelformat=simple]{subcaption}

\usepackage{booktabs,dcolumn}
\usepackage{multirow}
\usepackage{array}
\newcolumntype{P}[1]{>{\centering\arraybackslash}p{#1}}
\newcolumntype{M}[1]{>{\centering\arraybackslash}m{#1}}
\usepackage{diagbox}
\usepackage{dblfloatfix}
\usepackage{breqn}
\usepackage{bm}
\usepackage{tabu}
\usepackage{ulem}

\newcommand{\Tref}[1]{Table~\ref{#1}}
\newcommand{\eref}[1]{Eq.~(\ref{#1})}

\newcommand{\fref}[1]{Fig.~\ref{#1}}

\newcommand{\etal}{et al.~}
\newcommand{\eg}{e.g.~}

\renewcommand{\paragraph}[1]{\noindent\textbf{#1.}~}

\usepackage[pagebackref=true,breaklinks=true,letterpaper=true,colorlinks,bookmarks=false]{hyperref}

\begin{document}
% \renewcommand\thelinenumber{\color[rgb]{0.2,0.5,0.8}\normalfont\sffamily\scriptsize\arabic{linenumber}\color[rgb]{0,0,0}}
% \renewcommand\makeLineNumber {\hss\thelinenumber\ \hspace{6mm} \rlap{\hskip\textwidth\ \hspace{6.5mm}\thelinenumber}}
% \linenumbers
\pagestyle{headings}
\mainmatter
\def\ECCVSubNumber{1232}  % Insert your submission number here

\title{Deep Space-Time Video Upsampling Networks} % Replace with your title

% INITIAL SUBMISSION 
\begin{comment}
\titlerunning{ECCV-20 submission ID \ECCVSubNumber} 
\authorrunning{ECCV-20 submission ID \ECCVSubNumber} 
\author{Anonymous ECCV submission}
\institute{Paper ID \ECCVSubNumber}
\end{comment}
%******************

% CAMERA READY SUBMISSION
%\begin{comment}
\titlerunning{Deep Space-Time Video Upsampling Networks}
% If the paper title is too long for the running head, you can set
% an abbreviated paper title here
%
\author{Jaeyeon Kang\inst{1} \and
Younghyun Jo\inst{1}\and
Seoung Wug Oh\inst{1}
\and
\\
Peter Vajda\inst{2} \and Seon Joo Kim\inst{1,2}
}
\authorrunning{Kang et al.}
% First names are abbreviated in the running head.
% If there are more than two authors, 'et al.' is used.
%

%\and
%Springer Heidelberg, Tiergartenstr. 17, 69121 Heidelberg, %Germany
%\email{lncs@springer.com}\\
%\url{http://www.springer.com/gp/computer-science/lncs} \and
%ABC Institute, Rupert-Karls-University Heidelberg, Heidelberg, %Germany\\
%\email{\{abc,lncs\}@uni-heidelberg.de}}
\institute{\textsuperscript{1}Yonsei University, \textsuperscript{2}Facebook}
%\end{comment}
%******************
\maketitle

\begin{abstract}
Video super-resolution (VSR) and frame interpolation (FI) are traditional computer vision problems, and the performance have been improving by incorporating deep learning recently.  In this paper, we investigate the problem of jointly upsampling videos both in space and time, which is becoming more important with advances in display systems. One solution for this is to run VSR and FI, one by one, independently. This is highly inefficient as heavy deep neural networks (DNN) are involved in each solution. To this end, we propose an end-to-end DNN framework for the space-time video upsampling by efficiently merging VSR and FI into a joint framework. In our framework, a novel weighting scheme is proposed to fuse all input frames effectively without explicit motion compensation for efficient processing of videos. The results show better results both quantitatively and qualitatively, while reducing the computation time ($\times$7 faster) and the number of parameters (30\%) compared to baselines.
Our source code is available at \url{https://github.com/JaeYeonKang/STVUN-Pytorch}.
\keywords{Video Super-Resolution, Video Frame Interpolation, Joint space-time upsampling}
\end{abstract}

\section{Introduction}

In this paper, we introduce a method of upsampling both the spatial resolution and the frame rate of a video simultaneously.
This is an important problem as more high-performance TV displays are being introduced with higher resolution and frame rate, but the video contents have not yet caught up with the capabilities of displays.
For example, new UHD displays now come with 4K or even 8K resolution, and the frame rate of 120 fps. 
On the other hand, most available contents are still HD (1080p) or less in resolution, with the frame rate of 30 fps. 
Another potential application of this problem is the video replay for sports and security videos.
In order to inspect a video in much detail, videos are spatially magnified in slow motion. 
As shown by these examples, there is definitely a major need for a framework that can convert a given video into a video with higher resolution and frame rate.

\begin{figure}[t]
\small
\centering
\begin{subfigure}[b]{0.45\linewidth}
    \centering
    \includegraphics[width=\linewidth]{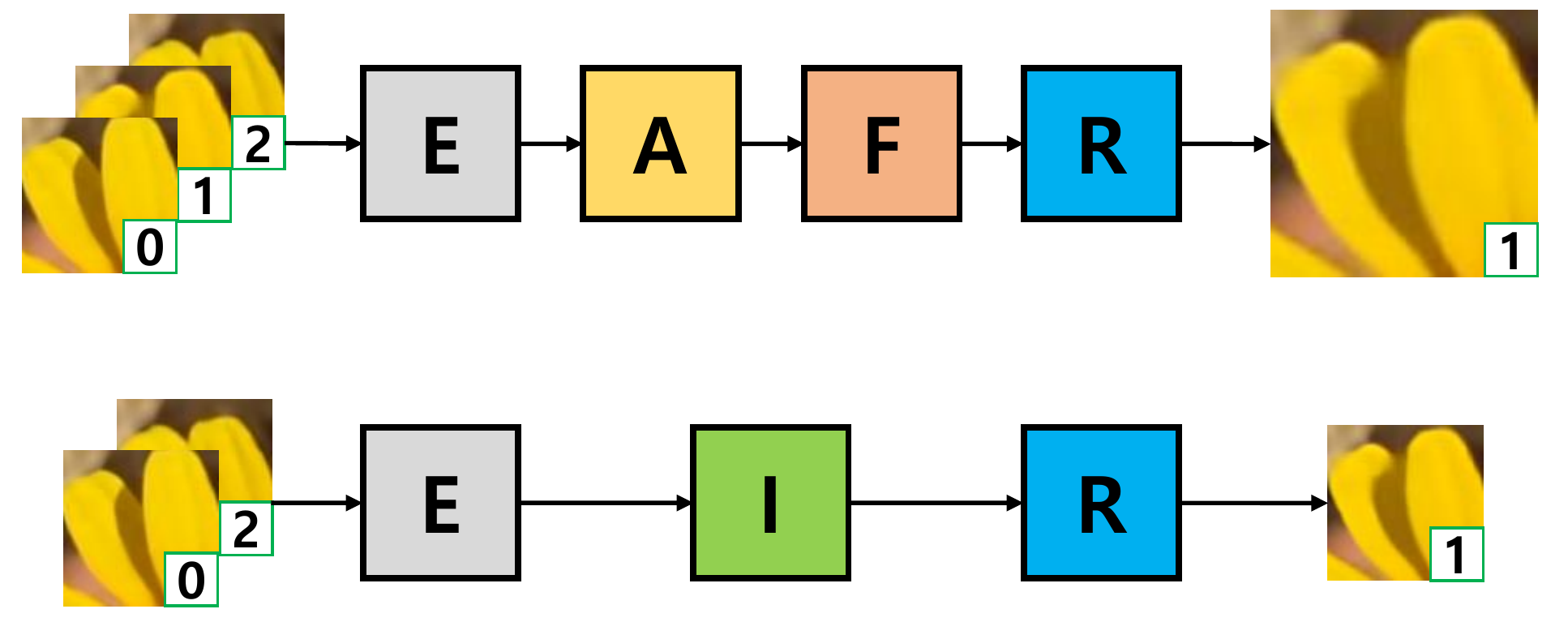}
    \caption{VSR and FI network.}
    \label{fig:Teaser1}
\end{subfigure}
\hfill
\begin{subfigure}[b]{0.53\linewidth}
    \centering
    \includegraphics[width=\linewidth]{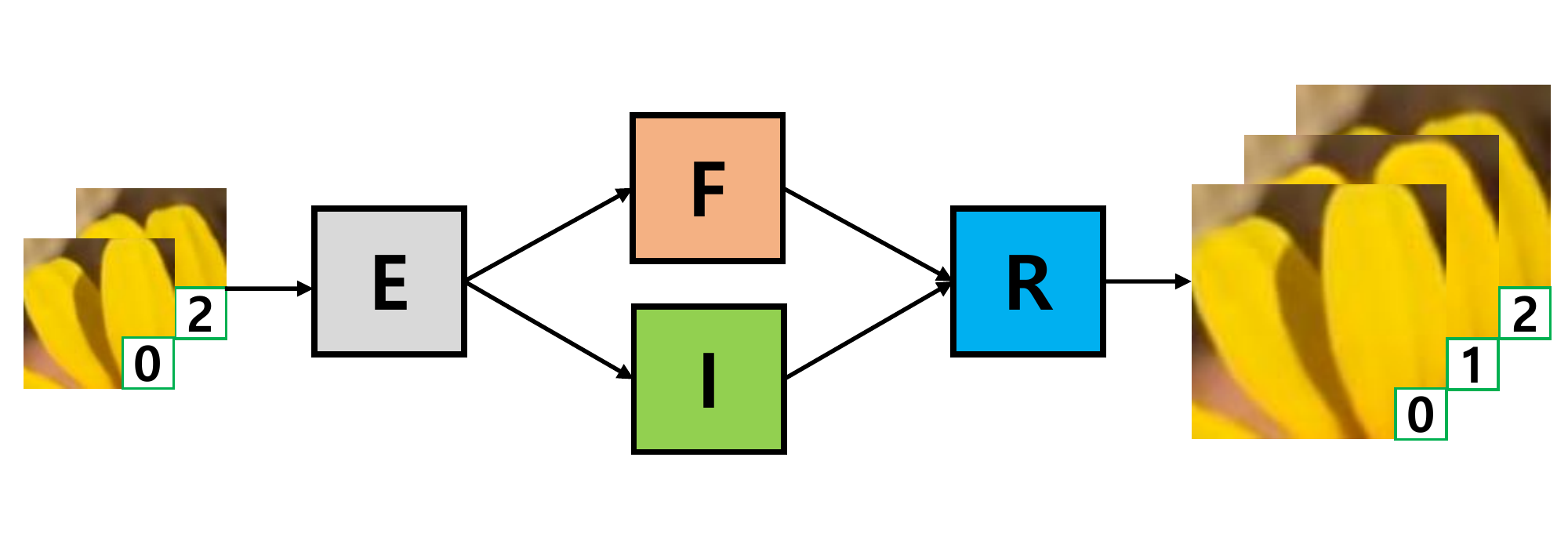}
    % \vspace{0.5pt}
    \caption{Our proposed network.}
    \label{fig:Teaser2}
\end{subfigure}
\caption{ 
Common pipelines for VSR/FI and the design scheme for our proposed network.
By efficiently merging the pipelines for the two tasks with new mechanisms for feature fusion and interactions between modules, we can accurately upsample videos both in space and time in a very efficient fashion.  E: extraction, A: alignment, F: fusion, R: reconstruction, I: interpolation.
}
\label{fig:Teaser}
% \vspace{-10pt}
\end{figure}

Deep neural networks (DNN) have become common solutions for video super-resolution (VSR) and frame interpolation (FI) recently.
With DNN, an obvious solution for the joint upsampling problem would be to sequentially run a VSR network followed by a FI network or vice versa.
However, running the two algorithms independently is computationally expensive and inefficient, as the state-of-the-art methods for each task employ heavy DNNs.
The goal of this paper is to design an efficient DNN for the joint space-time upsampling problem by investigating shareable components between the spatial and the temporal upsamping tasks. 

While there are many different DNN architecture for VSR~\cite{Tao_2017_ICCV,xue17toflow,sajjadi2018frame,wang2019edvr} and FI \cite{xue17toflow,jiang2018super,bao2019depth,niklaus2018context}, the common design schemes can be summarized as in \fref{fig:Teaser1}.
In VSR, most methods employ four stages -- feature extraction, alignment, fusion and reconstruction.
For FI, the process can be divided into feature extraction, feature interpolation and reconstruction. 

To jointly upsample videos both in space and time, we propose to combine the two tasks in an efficient manner as shown in \fref{fig:Teaser2} by sharing the common modules in feature extraction and reconstruction.
The modules are designed to interact and learn simultaneously for accurate and efficient reconstruction of jointly upsampled videos. 

Furthermore, we propose a novel way to efficiently fuse the features of individual frames for VSR without explicit motion compensation.
Most VSR methods rely on aligning many input frames through optical flows~\cite{xue17toflow,sajjadi2018frame,haris2019recurrent,Tao_2017_ICCV} or deformable convolutions~\cite{wang2019edvr} before fusing the extracted features.
As many methods use up to 7 input frames, aligning that many frames takes up a large portion of the computation.
To remove the computational burden of motion compensation, we fuse the feature maps without explicit alignment step (\fref{fig:Teaser2}).
In our feature fusion process, we propose Early Fusion with Spatio-Temporal weights (EFST) module that learns to fuse information by considering spatio-temporal relationship between input frames in an implicit manner. 
In this module, learnable spatio-temporal weights are computed in order to combine rich information from all frames instead of focusing too much on the target frame. 

There are no public datasets available for the joint space-time upsampling, as it is a relatively a new topic. 
While there are many datasets for VSR and FI separately, they are not ideal for the joint upsampling task.
To this end, we collected a new dataset called the Space-Time Video Test (STVT) dataset that can be used to evaluate joint upsampling methods. 
This dataset will be publicly available. 

In summary, the main contributions of our paper are as follows:
%In summary, our contribution can be summarized as follows:
% \begin{itemize}[leftmargin=*]
\begin{itemize}[noitemsep,topsep=0pt,leftmargin=*]
    \item By efficiently merging two networks of VSR and FI, we propose a novel framework called the Space-Time Video Upsampling Networks (STVUN) for joint space-time video upsampling. 
    With careful design of each module and their interactions, we produce better results while reducing the computation time ($\times 7$ faster) and the number of parameters (30\%) compared to sequentially connected state-of-the-art VSR and FI networks.
%    Also, our methods is very challenging as it can generate space $\times 4$ and time $\times \infty$ given per pixel in the input frame.
   
    \item We propose Early Fusion with Spatio-Temporal weights (EFST) to fuse input features efficiently without explicit motion compensation for VSR.
    
    \item Our framework can deal with more challenging upsampling tasks as it can upsample $4 \times 4$ in space and $\times \infty$ in time.
    In comparison, recent works on joint upsampling have only shown results on doubling the resolution both in space and time ($2 \times 2 \times 2$).
    % In our experiments, we show results for ~~~~~. 
    
    \item We collected Space-Time Video Test (STVT) dataset for evaluating the joint space-time upsampling task. This can be very useful for future work in this topic. 
\end{itemize}

\section{Related Work}
\subsection{Video Super-Resolution}

After Dong \etal \cite{dong2014learning} have successfully achieved the high performance by incorporating deep learning into the single image SR task, deep learning approaches  have also become prevalent in solving the VSR problem \cite{liao2015video,kappeler2016video,Caballero_2017_CVPR,Liu_2017_ICCV,Tao_2017_ICCV,sajjadi2018frame}.

DUF \cite{jo2018deep} used dynamic up-sampling filters to improve the resolution while reducing the flickering artifact which is prevalent in VSR task.
Their method takes the advantage of the implicit motion computed within the network, and additionally used the learned residual image to enhance the sharpness. 
RBPN \cite{haris2019recurrent} used an iterative refinement framework, which forwards the input frame with other frames at multiple times. 
They use the idea of back-projection, which computes a residual image for each time step to reduce the error between the target and the output. 
In EDVR \cite{wang2019edvr}, input frames are first aligned with the target frame using the deformable convolution~\cite{dai2017deformable}.
Aligned frames are then fused using the temporal and spatial attention (TSA) mechanism. 
%We modify TSA for other purpose and use to fuse unaligned reference frames.
%%
%% 여기서 차이점을 좀 언급해주는 게 좋을 것 같은데
%%

\subsection{Video Frame Interpolation}
Video frame interpolation can be roughly divided into two categories: kernel-based methods and optical flow-based methods.
As an interpolation kernel based approach, Niklaus \etal \cite{niklaus2017videoada} proposed AdaConv, which produces interpolation kernels to generate intermediate frame.
In \cite{niklaus2017video}, they extended the method to reduce the computational cost, which is named as SepConv using 1D kernels instead of 2D kernels.

With the introduction of CNN-based optical flow algorithms~\cite{fischer2015flownet}, several frame interpolation algorithms using the optical flow have been developed.
%After FlowNet \cite{fischer2015flownet} was introduced to learn optical flow %through CNN, several methods using the optical flow have been developed.
Liu \etal \cite{liu2017video} produce intermediate frames by the trilinear sampling based on the estimated deep voxel flows called DVF.
%predicts a voxel flow, not only for the spatial axis but also for the %temporal axis, and this can produce intermediate frame by the trilinear sampling from two input frames.
Xue \etal \cite{xue2019video} used the bi-directional flow to warp both input frames using the backward warping function.
Jiang \etal \cite{jiang2018super} obtained the bi-directional flow through the network and then linearly transformed two flows with respect to the time value to generate multiple intermediate frames. 
Niklaus and Liu \cite{niklaus2018context} used the forward warping and further designed a refinement network in order to fill the holes caused by the forward warping.
Liu \etal \cite{liu2019deep} used cycle consistency loss to enhance synthesized frames to be more reliable as input frames. 
To deal with the occlusion problem which is a common issue in optical flows, additional depth information was used to refine the optical flows in DAIN~\cite{bao2019depth}.

\subsection{Space-Time Upsampling}

% There are only few works that have covered the joint upsampling of videos both in space and time without using DNN. 
In \cite{shechtman2005space}, Shechtman \etal  first proposed a space-time super-resolution framework by using multiple low resolution (LR) videos of the same dynamic scene.
Different from the frame interpolation methods mentioned above, they explicitly deal with the motion blur to generate sharp interpolated frames.
In \cite{shahar2011space}, Shahar \etal extended the work in \cite{shechtman2005space} with a method that only uses a single video to enhance the resolution. 
Sharma \etal \cite{sharma2017space} first used a DNN architecture for the joint space-time upsampling. 
They used the auto-encoder to learn the mapping between LR and high resolution (HR) frames, and the frame interpolation was simply done by the tri-cubic interpolation. 
% Recently, ~~~ \textcolor{red}{please add more description on FISR here.}
%Recently, with the enhancement of the %resolution and frame rate in TV displays, 
Another deep joint upsampling method called
FISR \cite{kim2020fisr} was recently introduced, which targets for estimating 4K, 60fps video from 2K, 30fps video.
% They used multiple consecutive frames per each iteration and the outputs of each consecutive frames are used to regularize their joint network with multiple temporal losses.
% They trained the joint upsampling network by giving temporal losses to the multiple outputs that put several consecutive frames in an iteration.
They regularized their joint upsampling network by forwarding multiple chunks of frames into one iteration and set multiple temporal losses at the output of each chunk. 
% FISR \cite{kim2020fisr} use multiple temporal losses to regularize for space-time upsampling network.
% generate sharp results with removing flickering artifact.
Note that, FISR only generates 8 pixels (space $\times2$, time $\times 2$) per input pixel, while our work aims at more challenging task of generating more pixels (e.g. 64 pixels for space $\times4$, time $\times 4$).%, which is more challenging.
% As it is trained for targeting 4K resolution, FISR only can generate 4 pixels (space $\times2$, time $\times 2$) per input pixel, while our work aims to generate more pixels (space $\times4$, time $\times \infty$), which is more challenging. 

% FISR \cite{kim2020fisr} use multiple temporal losses to generate sharp results with removing flickering artifact.
% FISR aims at generating 4 pixels (space $\times2$, time $\times 2$) per input pixel, while our work aims to generate more pixels (space $\times4$, time $\times \infty$), which is more challenging. 

\section{Space-time Video Upsampling Algorithm}
Given a sequence of LR frames $X_t$, our method produces HR frames $\hat{Y}_t$ of inputs as well as the intermediate HR frames $\hat{Y}_T$ in-between the input frames.
The term $t$ denotes the input time index, and $T$ indicates the newly created time index.
The size of a LR frame is $H \times W \times C$, where $H$, $W$, and $C$ are the height, the width, and the number of channels respectively.
The output size is $rH \times rW \times C$, with $r$ being the spatial upscaling factor.
We can generate $N$ multiple upsampled intermediate frames in-between the two input frames.
The problem is very challenging as the algorithm has to generate $r^2 \times (N+1)$ pixels per pixel in the input frame. For example, we need to generate 64 pixels in the output per input pixel with $r=4$ and $N=3$.

\begin{figure}[t]
% 	\vspace{5pt}
	%\centering
	\begin{center}
		\includegraphics[width=1\linewidth]{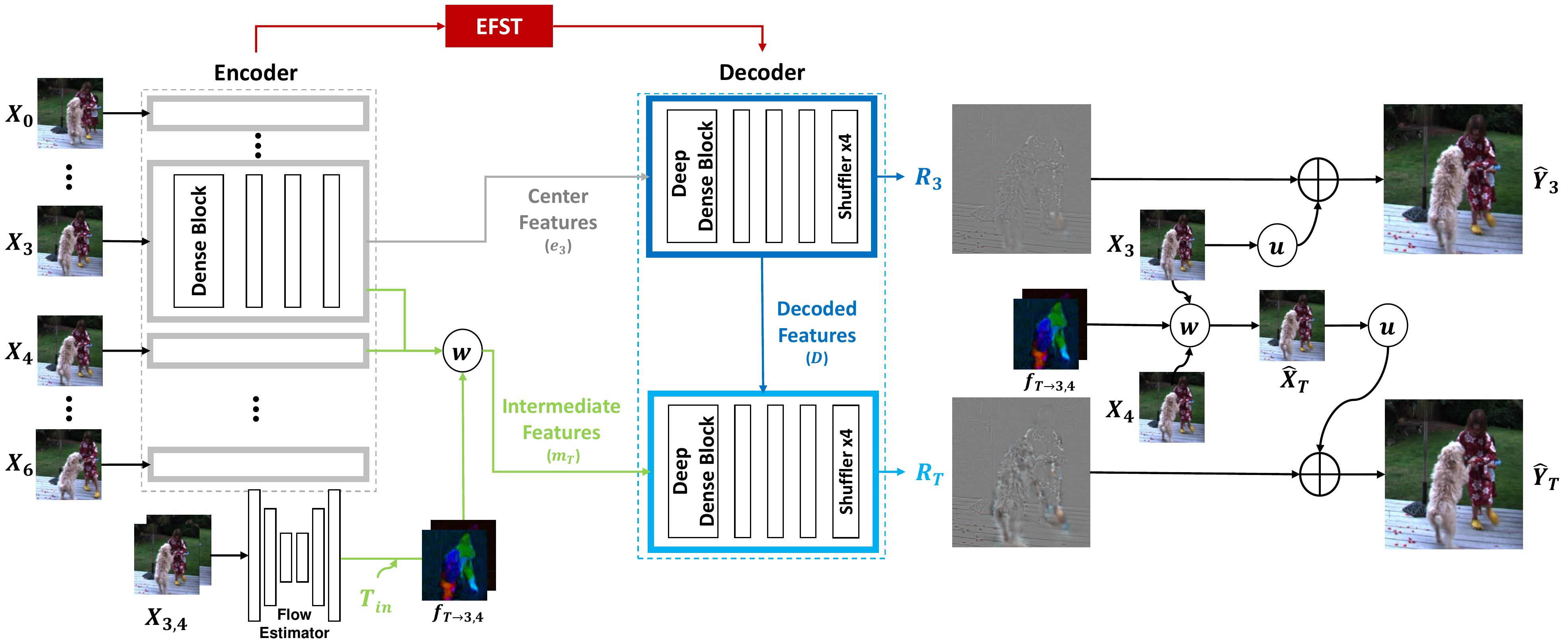}
	\end{center}
	\vspace{-15pt}
	\caption{Overview of our space-time upsampling network. 
	The network is composed of several interacting modules to produce the HR frames of the center frame $\hat{Y}_3$ and the intermediate frames $\hat{Y}_T$, where $T \in{[3,4]}$.
%	Space upsampling in dark blue and space-time upsampling in sky share same weights.
	The term $w$ and $u$ indicate backward warping and bilinear upsampling respectively.}
	\label{fig:Network}
% 	\vspace{-20pt}
\end{figure}

\subsection{Network Overview}
The overview of our network is shown in \fref{fig:Network}.
Our network is composed of multiple modules: encoder, feature fusion for spatial upsampling (EFST), flow estimator for frame interpolation, and decoder.
Our framework takes 7 LR frames as inputs, for example $X_{[0,1,2,3,4,5,6]}$.
Then, it produces the HR frames for the center frame $\hat{Y}_{3}$ as well as $N$ HR intermediate frames $\hat{Y}_{T}$, where $T \in{[3,4]}$.

The encoders that share weights are first used to extract features per frame.
The encoded features are fused using EFST for the spatial upsampling, and interpolated using the computed flows for the temporal upsampling. 
The decoding block that consists of decoders with shared weights produces residual images for the spatial and the temporal upsampling, both of which are added to the bilinearly upsampled images to produce the final output frames.  

\subsection{Network Details}

\subsubsection{Encoder}
\begin{figure}[t]
% 	\vspace{5pt}
	%\centering
	\begin{center}
		\includegraphics[width=1\linewidth]{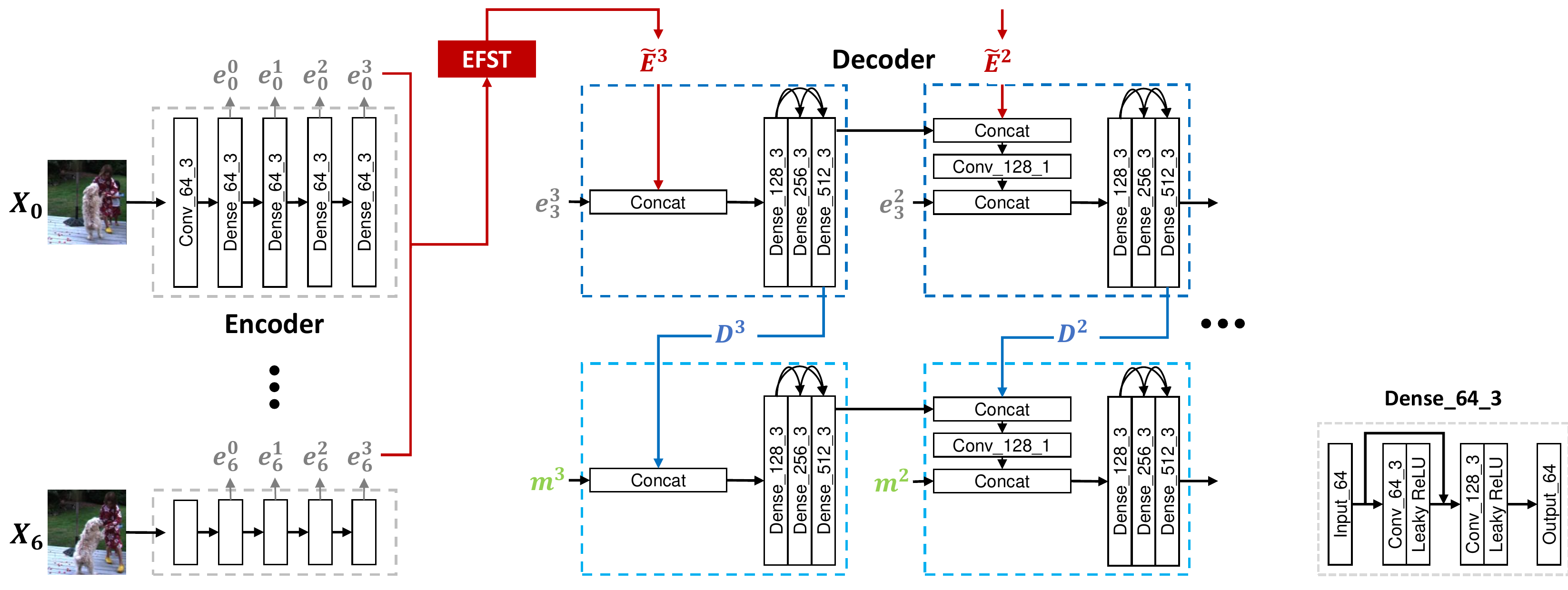}
	\end{center}
	\vspace{-15pt}
	\caption{The structure of the encoder and decoder. 
	Dense\_64\_3 denotes dense block with 64 input channel dimension and $3\times3$ kernel size.
	We use Leaky ReLU with the slope value 0.1.}
	\label{fig:Encoder_Decoder}
	\vspace{-5pt}
\end{figure}

Structure of the encoder is shown in \fref{fig:Encoder_Decoder}.
The encoder extracts feature representations for each frame and consists of multiple dense convolution blocks.
Each dense block is connected to the corresponding block in decoder through EFST.
This allows the decoder to keep considering the temporal relationship of inputs.
The encoded features are expressed as $e^{i}_{t}$, where $i$ is the block index and $t$ is the time index of the input frame.

\subsubsection{Early Fusion with Spatio-Temporal weights (EFST)}
In most VSR methods, features from multiple frames are aligned before fusion using explicit motion compensation by optical flows or deformable convolutions.
However, aligning multiple frames (6 in most cases including ours) to the center frame is computationally expensive. 
Therefore, we exclude explicit alignment process by devising the EFST module for implicit feature alignment and fusion for spatial upsampling. 

To merge input features from the encoders, we first apply early fusion to reduce the computational cost.
Early fused features $E^i$ are defined as $Conv(Concat$
$[e^{i}_{0}, ... ,e^{i}_{6}])$ where $Conv$ reduces the channel dimension by the factor of 7 with an $1\times1$ convolution filter.
% Early fused features $E^i$ can be generated by concatenating all the 7 encoded input features $Concat[{e^{i}_{0}, ... ,e^{i}_{6}}]$ and then reducing the channel dimension by the factor of 7 with an $1\times1$ convolution filter.
However, since the early fusion will collapse all temporal information in the first layer, the features of the target (center) frame will be mainly used as mentioned in \cite{shi2016real}.
% In addition, information in other frames may vanish due to the bottleneck.
Some information in other frames may vanish due to the bottleneck.
This is because most information for reconstructing the HR target frame is contained in the input LR center frame.

\begin{figure}[t]
% 	\vspace{5pt}
	%\centering
	\begin{center}
		\includegraphics[width=0.85\linewidth]{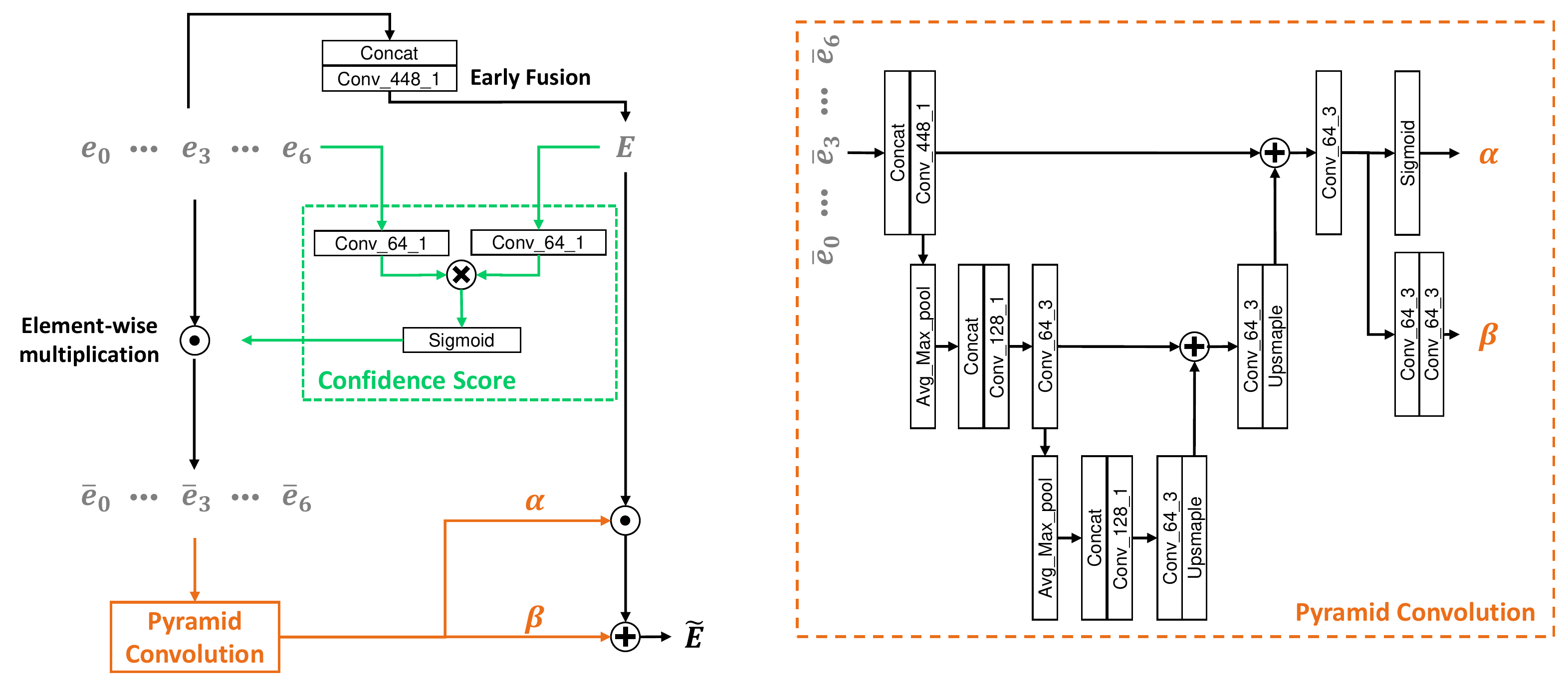}
	\end{center}
	\vspace{-15pt}
	\caption{Early Fusion with Spatio-Temporal weights (EFST) module.
% 	The confidence score is computed from early fusion result $E^i$ for temporal attention.
    Avg\_Max\_pool means pooling separately with average pooling and max pooling.
	For temporally attending more frames, the confidence score is computed from early fusion result $E^i$ and $e^{i}_{t}$.
	Then, the spatio-temporal weights $\alpha$ and $\beta$ are computed and applied.
	For simplicity, we omit the superscript $i$, and see text for details.}
	\label{fig:EFST}
	\vspace{-5pt}
\end{figure}

In order to use valuable information in the features from all the input frames without explicit alignment, we propose the EFST module that computes spatio-temporal weights to compensate $E^i$.
% to combine input feature maps.
The structure of EFST is shown in \fref{fig:EFST}.
In the early fusion result $E^i$, most input features are not considered equally since $E^i$ will be computed to mainly focus on the center frame.
Therefore, we design a confidence score to effectively fuse informative features from the neighbor frames as well as the center frame.
We estimate the confidence score by computing dot-product between $E^i$ and all the $e^{i}_{t}$.
We use this confidence score as a temporal attention to find which frames need to be more referred.
The confidence score is computed as follows: 
\begin{equation}
    \begin{aligned}
        s^{i}_{t} = \text{sigmoid}(\theta(e^{i}_{t})\circ \delta(E^i)),
    \end{aligned}
    \label{eq:Dissimilarity}
\end{equation}
where $\circ$ is dot-product and $s$ is the confidence score.
% $\theta$ and $\delta$ are single convolutional layer with filter size $1\times1$ to align the same feature space.
$\theta$ and $\delta$ are single convolutional layer with filter size $1\times1$.
$s^{i}_{t}$ has the same spatial size as $e^{i}_{t}$ and the values of $s^{i}_{t}$ are in $[0,1]$.
To pay more attention to the frames with high confidence score, we multiply this value to the original encoded features $e^{i}_{t}$ as follows:
\begin{equation}
    \begin{aligned}
        \bar{e_t}^{i} = s^{i}_{t} \odot e^{i}_{t},
    \end{aligned}
    \label{eq:Temporal attention}
\end{equation}
where $\odot$ denote element-wise multiplication.

% After that, all temporally weighted encoded features $\bar{e_t}^{i}$ are concatenated and forwarded to several convolution filters with pyramid design.
All temporally weighted encoded features $\bar{e_t}^{i}$ are then concatenated and forwarded to pyramid designed convolutional layers to further consider spatio-temporal information.
Pyramid convolution can effectively enlarge the receptive field with just few convolution layers.
Afterwards, we generate learnable spatio-temporal weights $\alpha, \beta$.
It is a tensor with same size of $E^i$.
It transform the initial early fusion result to learn the alignment in an implicit way.
The final fused features is computed as follows:
% The final fused features can be obtained as follows:
\begin{equation}
    \begin{aligned}
        \tilde{E}^{i} = \alpha \odot E^i + \beta.
    \end{aligned}
    \label{eq:EFST}
\end{equation}
% 

% The overall structure is similar with TSA~\cite{wang2019edvr}.
Our EFST module is similar to Fusion with Temporal and Spatial Attention called TSA in \cite{wang2019edvr}.
% TSA uses similarity distance to focus more on well-aligned frames, since misalignment can severely interfere with learning.
TSA measures similarity distance between aligned frames and target frame to temporally weight more on well-aligned frames, since misalignment can severely interfere with learning.
In comparison, we use confidence score as a way to involve more features from more input frames, which eventually works as a joint alignment and fusion process without explicit alignment.

\subsubsection{Intermediate Feature Interpolation}
The feature interpolation process is shown in green in \fref{fig:Network}. 
Features $m^{i}_{T}$ of an intermediate frame are synthesized by warping the features of the input frames ($X_{3},X_{4}$) using the optical flow estimated by the flow estimator. 
To reduce the computational cost, we warp the encoded features of the two inputs to produce the target intermediate features instead of first creating the intermediate frame and encoding it again.
Note that explicit alignment is only used for intermediate frames but not for merging all input features (EFST).
We first use the optical flow computed by the PWC-Net \cite{sun2018pwc}.
Then, we use the following formulation \cite{jiang2018super} to estimate the flow between the input frames and the intermediate frame: 
\begin{equation}
    \begin{aligned}
        f_{T\rightarrow{3}} = -(1-T_{in})T_{in}f_{3\rightarrow{4}} + T_{in}^2f_{4\rightarrow{3}}, \\
        f_{T\rightarrow{4}} = (1-T_{in})^2f_{3\rightarrow{4}}  -T_{in}(1-T_{in})f_{4\rightarrow{3}},
    \end{aligned}
    \label{eq:OpticalFlow}
\end{equation}
where $f$ indicates the optical flow and $T_{in}$ is a relative scalar value (\eg when we want to get $\hat{Y}_{3.5}$, then $T_{in}$ is set to 0.5).
Note that $T_{in}$ is given as an input to the network to provide the time index of the intermediate frame.

Finally, the features of the intermediate frame are computed as follows: 
\begin{equation}
    \begin{aligned}
        m^{i}_{T} = {w(e^{i}_{3}, f_{T\rightarrow{3}}) + w(e^{i}_{4}, f_{T\rightarrow{4}})\over {2}},
    \end{aligned}
    \label{eq:IntermediateFeatures}
\end{equation}
where $w$ stands for the backward warping. 

At the same time, we generate LR intermediate frames $X_T$ for the subsequent process of bilinear upsampling with
\begin{equation}
\begin{aligned}
\hat{X}_{T} = {w(X_{3},f_{T\rightarrow{}3}) + w(X_4,f_{T\rightarrow{}4}) \over {2}},
\end{aligned}
\label{eq:IntermediateBlurredFrame}
\vspace{2pt}
\end{equation}
and for finetuning PWC-net with ground-truth intermediate frames during training whole network, we set loss function as:
\begin{equation}	
\mathcal{L_M} = \sum_{T}||\hat{X}_{T} - {X}_{T}||_1. 
\end{equation}
% Note that we do use alignment to synthesize intermediate frames but not for merging all encoded features for VSR.

\subsubsection{Decoder}

The decoder reconstructs target HR residual image and it consists of multiple dense convolution blocks.
The same number of blocks is used to connect with each block of the encoder.
We design a more deeper dense block which is shown in \fref{fig:Encoder_Decoder}, since more layers and connections could boost performance \cite{lim2017enhanced,wang2018esrgan}.
To generate the residual image of the target frame, features from the last layer of the last block are convolved with a filter having $C\cdot r\cdot r$ output channels, where the output is then reshaped to the size of $rH \times rW \times C$ through the pixel shuffler \cite{shi2016real} with the scale factor of 4.

For space upsampling, features of target (center) frame $e_3$ and fused features from EFST $\tilde{E}$ are used as inputs.
Output residual image $R_{3}$ is added to the upsampled target frame to generate final HR output as follows:
%%%
\begin{equation}
    \begin{aligned}
        \hat{Y}_{3} = u(X_{3})+R_{3},
    \end{aligned}
    % \vspace{2pt}
\end{equation}
where $u(\cdot)$ is the bilinear upsampling function.
The loss function for space upsampling is defined as:
\begin{equation}	
\mathcal{L_S} = ||\hat{Y}_{3} - Y_{3} ||_1.
\end{equation}

Since the decoder consists of more deeply stacked convolution layers, it creates more refined spatio-temporal information from the EFST features.
Therefore, during the space upsampling task, the decoded features $D^i$ are generated at the end of each dense block and forwarded to the space-time upsampling task to supplement with more rich information.
For space-time upsampling, the intermediate features $m_T$ are passed to another decoder that shares weights. % with the decoder for space upsampling.
Here, different from space upsampling, decoded features $D^i$ are fed instead of feeding $\tilde{E}$. 

Then, HR residual image $R_T$ of the intermediate frame is generated and the final HR intermediate frame is computed as follows:
\begin{equation}
    \begin{aligned}
        \hat{Y}_T = u(\hat{X}_T)+R_T,
    \end{aligned}
\end{equation}
where $\hat{X}_T$ is from \eref{eq:IntermediateBlurredFrame}. 
Our loss function to train space-time upsampling is:
% As we do not have the access to the ground-truth optical flows, we have to use the final warped results to define the loss for the temporal decoder:
\begin{equation}	
\mathcal{L_F} = \sum_{T}||\hat{Y}_{T} - Y_T ||_1,
\end{equation}
where $T$ can be any values in $[3,4]$. 
Note that we can generate arbitrary number of intermediate frames using \eref{eq:OpticalFlow}.

\subsection{Training}
% Why we do not use Vimeo dataset ...
Vimeo septuplets dataset \cite{xue2019video} is usually used to train VSR and FI tasks.
But the length of video frames in Vimeo dataset is too short for our task.
It consists of 7 frames per clip, but we need at least 8 frames for training.
%  with intermediate frames
Therefore, we collect training videos of 240fps from YouTube.
This training dataset consists of various scenes with global camera motions and local object motions.
In total, the dataset contains about 1800 video clips and 220K frames.
To make LR frames, HR frames are first smoothed with a Gaussian filter and then subsampled with respect to the scaling factor $r=4$.
For the data augmentation, we randomly flip left-right and rotate 90/180 degrees. 
We also reverse the order of the sequence to enlarge the training dataset.
The whole training and test is processed in RGB channels.

It is difficult to train all the networks in our framework simultaneously from scratch, as there are many interactions between the components.
We first pretrain the encoder and the spatial decoder by minimizing $\mathcal{L_{S}}$ (VSR part only).
% Note that this combination of modules works as a VSR network and we name them as Space Video Upsampling Networks (SVUN). 
For this pretraining, we use 7 frames $Y_{[0,1,2,3,4,5,6]}$ in the training dataset and $128 \times 128$ patches are cropped. 
% for the training. 
We use the Adam optimizer for 300K iterations with the mini-batch size of 32.
The learning rate is initialized to 0.0001 and decreased by a factor of 2 every 100K iterations.

%  named Space-time Video Upsampling Networks (STVUN)
After pretraining the VSR part, we train the whole network using the following total loss function:
\begin{equation}
    \mathcal{L} = \lambda_{\mathcal{M}}\mathcal{L_{M}} + \lambda_{\mathcal{S}}\mathcal{L_S} + \lambda_{\mathcal{F}}\mathcal{L_F},
\end{equation}
where $\lambda_{\mathcal{M}}$, $\lambda_{\mathcal{S}}$, and $\lambda_{\mathcal{F}}$ are the weight parameters.
In our experiment, we empirically set $\lambda_{\mathcal{M}}=1, \lambda_{\mathcal{S}}=1$, and $\lambda_{\mathcal{F}}=1$ for the best results.
For the joint training, $256\times256$ patches are used rather than $128\times128$ in order to deal with large motions.
Intermediate frames in-between $Y_3, Y_4$ (\eg $Y_{3.5}$) as well as the 7 frames are used for training VSR and FI part together.
We train the whole network for 400K iterations and the initial learning rate is set to 0.00005.
The same learning rate decay is used.

\section{Experiments}
In this section, we provide both quantitative and qualitative evaluations of our algorithm. 

\subsubsection{Testsets}
% As joint space-time upsampling is a relatively a new topic, there are not many datasets that we could use for both tasks.
While there are some datasets for VSR and FI separately, they are not ideal for the joint space-time upsampling task. 
For example, the Vid4 testset \cite{liu2014bayesian} for VSR have a lot of details, but the motion between the frames is too small. 
This limits the assessment of FI performance.
MPI Sintel testset \cite{Butler:ECCV:2012} is synthetic dataset which dose not have much detail to assess VSR performance.
% REDS dataset~\cite{nah2019ntire} is recently used for both VSR and video deblurring tasks.
% However, this dataset is not optimal for evaluating the space-time upsampling task as it contains unnatural camera movements for motion blur.
REDS-VTSR dataset~\cite{nah2019ntire} is used for VSR and FI separately, but it contains unnatural camera movements.
% In addition, we need at least 15 frames for each scene to evaluate the performance.
% For this reason, both Vimeo testset~\cite{xue17toflow} and Middlebrry testset~\cite{baker2011database} could not be used
In addition, the Vimeo ~\cite{xue17toflow}, Middleburry~\cite{baker2011database} and FISR \cite{kim2020fisr} testset are not available, since at least 15 frames are required for each scene to evaluate the performance.
% The testset of FISR \cite{kim2020fisr} also could not be used as the length of videos is 7 frames.
% The testset of FISR \cite{kim2020fisr} also could not be used as the length of videos is 7 frames.

% To this end, we use Vid4, MPI Sintel, REDS-VTSR~\cite{nah2019ntire} testset that are frequently used in VSR and FI respectively for testing the generalization our performance.
To this end, we use Vid4, MPI Sintel and REDS-VTSR~\cite{nah2019ntire} for testing the generalization our performance.
In addition, we create Space-Time Video Test (STVT) dataset that consists of 12 dynamic scenes with both natural motions and spatial details for the joint upsampling evaluation.
Each scene has at least 50 frames, and we will make STVT dataset publicly available to promote more research in this topic.

\subsubsection{Baselines}
We make two baseline methods ($V \rightarrow{F}$ and $F \rightarrow{V}$) that combine VSR and FI, which run consequently.
$V$ and $F$ indicate VSR and FI respectively.
For example, $F \rightarrow{} V$ indicates running FI first and then VSR.
For $V$ and $F$, we use EDVR~\cite{wang2019edvr} and DAIN~\cite{bao2019depth} respectively, which are the state-of-the-art methods with publicly available codes.
% DAIN~\cite{bao2019depth} is used for $F$, which is one of the state-of-the-art methods with publicly available codes.
As the bias of the dataset affect the evaluation performance \cite{tommasi2017deeper}, for fair comparison, we try to finetune the baseline methods with our YouTube training dataset.
However, since their weights are already highly finetuned, we find that the performance is rather reduced when we jointly train both networks at the same time (0.15dB is reduced for Vid4 testset).
Therefore, we fix their weights to produce the results.
We also compare our method with FISR~\cite{kim2020fisr}, the only deep learning based work that we can compared to at this moment.

\subsection{Comparisons}
For the evaluation, we extract odd numbered frames in the testset and set them as ground-truth frames. 
Only the even numbered frames are used to generate the space-time upsampled results.
% We set $T_{in}=1/2$ to compare our space-time upsampling results with two baseline methods.
We first compare our method with the two baseline methods.
We set $T_{in}= 1/2$ for generating the HR intermediate frame.
% We set $T_{in}=1/2$ to compare our space-time upsampling results with two baseline methods.
\Tref{tab:STVUN_quantitative} shows the quantitative results of different approaches for $\times4$ space and $\times2$ time.
In every testset, $F \rightarrow V$ consistently shows the worst performance, because FI works better on HR input frames due to sufficient details.
On the other hand, in the case of $V \rightarrow F$, FI can access sufficient details from VSR, thus it can generate sharper results.
However, the improvement in the resolution increases the amount of computation for FI ($\times 4$ slower).

%% Qunatitative evaluation 
\begin{table}[t]
\centering
\caption{Quantitative evaluation of the joint space-time upsampling on multiple testsets.
We compare our method with the two baseline approaches by measuring the PSNR and SSIM.
We set $T_{in} = 1/2$ for comparison.
We also write down the number of parameters and the running time for each method.
The running time is measured when generating the results with the resolution $960 \times 540$.
The best is shown in bold.
% Our method is shown in bold and shows the highest accuracy and efficiency.
}
\vspace{1pt}
	\begin{tabular}
		{@{}P{0.20\linewidth}||P{0.18\linewidth}|P{0.18\linewidth}@{}P{0.18\linewidth}@{}P{0.18\linewidth}}
		\hline
		& & $F \rightarrow{} V$ & $V \rightarrow{} F$ &  Ours\\
		\hline
		\hline
% 		\multirow{2}{*}{\vspace{-2pt}\shortstack[l]}
% 		Dataset & Vid4 & 25.22/0.7506 & 26.39/0.8163 & \textbf{26.52/0.8239} \\
% % 		& SPMCS & 30.74/0.8718 & 31.85/0.8919 & 31.65/0.8909 \\
% 		& Sintel & 26.99/0.7986 & \textbf{27.56/0.8185} & 27.55/0.8133 \\
% 		& STVT & 26.43/0.8435 & 26.96/0.8619 & \textbf{27.23/0.8642} \\
		Dataset & Vid4 & 25.22/0.7506 & 26.39/0.8163 & \textbf{26.49/0.8231} \\
% 		& SPMCS & 30.74/0.8718 & 31.85/0.8919 & 31.65/0.8909 \\
		& Sintel & 26.99/0.7986 & 27.56/\textbf{0.8185} & \textbf{27.58}/0.8134 \\
		& REDS-VTSR & 23.70/0.6541 & 23.63/0.6533 & \textbf{23.78}/\textbf{0.6601} \\
		& STVT & 26.43/0.8435 & 26.96/0.8619 & \textbf{27.23/0.8644} \\
		\hline
		\hline 
		\#Params & & 44.7M & 44.7M & \textbf{30.9M} \\
		Running Time &  & 0.52s & 2.14s & \textbf{0.30s} \\
		\hline
	\end{tabular}
	\vspace{-10pt}
\label{tab:STVUN_quantitative}
\end{table}

%% Compare with FISR
\begin{table}[t]
\centering
\caption{Comparison with FISR \cite{kim2020fisr}. 
We train our model with the upsampling factor space $\times 2$ and time $\times 2$ which is the same as FISR.
% The overall environment of experiment is the same as \Tref{tab:STVUN_quantitative}, and only the input size is twice bigger for generating outputs.
}
\vspace{1pt}
	\begin{tabular}
		{@{}P{0.20\linewidth}||P{0.18\linewidth}|P{0.18\linewidth}@{}P{0.18\linewidth}}
		\hline
		& & FISR~\cite{kim2020fisr}  &  Ours\\
		\hline
		\hline
		Dataset & Vid4 &26.93/0.8534 & \textbf{30.60/0.9369} \\
		& Sintel & 27.17/0.8115 & \textbf{28.36/0.8329} \\
		& REDS-VTSR & \textbf{23.89}/\textbf{0.6601} & 23.66/0.6550 \\
		& STVT & 26.49/0.8514 & \textbf{28.01/0.8895} \\
		\hline
		\hline 
		\#Params & & 62.3M & \textbf{30.9M}  \\
		Running Time & & 1.10s & \textbf{0.98s} \\
		\hline
	\end{tabular}
	\vspace{-15pt}
\label{tab:FISR}
\end{table}

Our results show better performance for all datasets as shown in \Tref{tab:STVUN_quantitative}.
The performance difference in Vid4, Sintel and REDS-VTSR testset is not that big because those testsets are not constructed for this particular tasks and not optimal for evaluating the joint upsampling task.
The performance gap between our method and the baselines become larger with the STVT dataset, which is specifically
designed for the joint upsampling.
%As mentioned before, Vid4 and Sintel testsets are not proper for evaluating the joint upsampling task.
% The performance gap in Sintel dataset seems minor.
%We believe this is why the performance gap in Sintel dataset is minor.
%In Vid4, our method generates better results than the other baseline methods.
%As STVT dataset is proper for evaluating joint upsampling task, the performance gap becomes much bigger.
% As can be seen from the results, our method generate considerable results than other baseline approaches.

\Tref{tab:STVUN_quantitative} also shows the number of parameters and the computation time of different methods.
In this experiment, our total parameters and computational times does include PWC-Net~\cite{sun2018pwc}.
We run the methods on Nvidia Geforce Titan X and measure the time taken to generate one $ 960 \times 540 $ jointly upsampled frame.
The number of parameters is reduced by more than 30\% compared to the baseline methods, and
% In addition, since the alignment stage is removed, ours is 5 times faster than $V \rightarrow F$ and 2.5 tmes faster than $F \rightarrow V$.
the speed is 7 times faster than $V \rightarrow F$ and 1.7 times faster than $F \rightarrow V$.
Although ours is lighter than the baseline methods, it exceeds the performance of baseline methods, indicating that our network is designed efficiently.

%%%%%%%%%%%%%%%%%%%%%%%%%%%%%%%%%%%%%%%%%%%%%%%%%%%%%%%%%%%%%

\begin{figure}
\scriptsize
\centering
% \begin{tabular}{@{}p{0.9\linewidth}@{}}
    \begin{subfigure}[h]{\linewidth}
    \centering
    \begin{tabular}
        {@{}m{0.25\linewidth}@{\hskip2pt}m{0.1\linewidth}@{\hskip2pt}m{0.12\linewidth}@{\hskip2pt}m{0.12\linewidth}@{\hskip2pt}m{0.12\linewidth}@{\hskip2pt}m{0.12\linewidth}@{\hskip2pt}m{0.12\linewidth}@{}}
        &
        &
        \centering~$\hat{Y}_{0}$ &
        \centering~$\hat{Y}_{0.25}$ &
        \centering~$\hat{Y}_{0.5}$ &
        \centering~$\hat{Y}_{0.75}$ &
        \centering~$\hat{Y}_{1}$ \tabularnewline
        
        \multirow{3}{*}{\shortstack{\includegraphics[width=\linewidth]{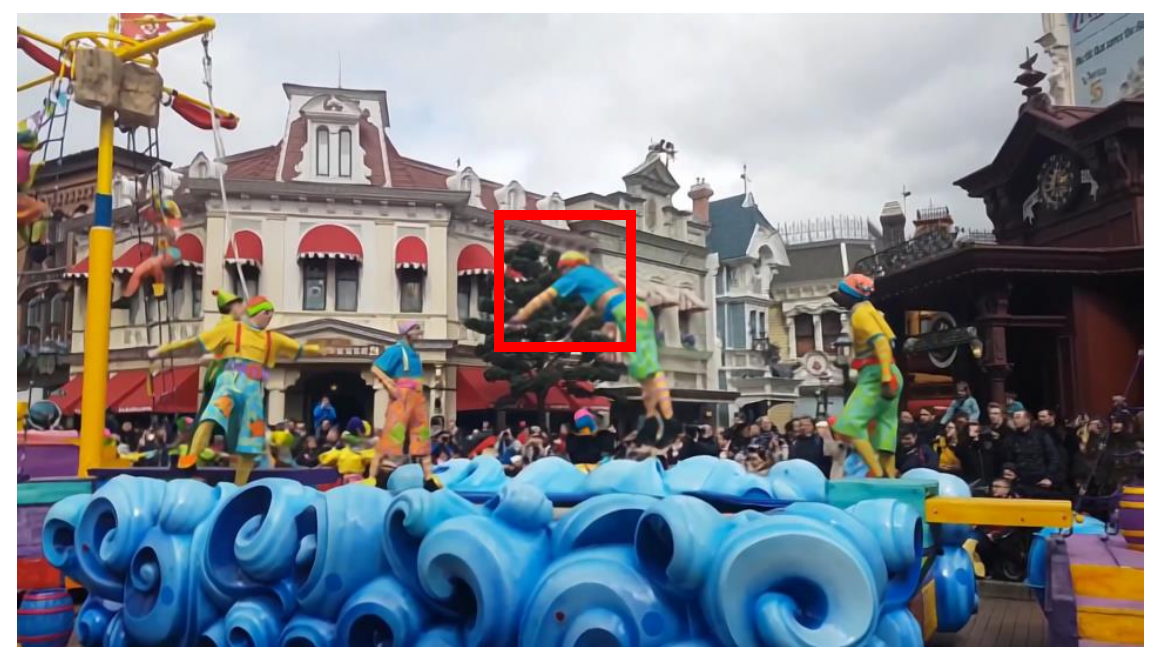}\\Parade\_1}} & 
        \multicolumn{1}{!{\vrule width 1pt}c}{\centering~$F \rightarrow V$} &
        \includegraphics[width=\linewidth]{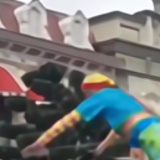} &
        \includegraphics[width=\linewidth]{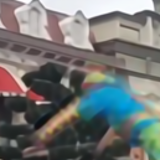} &
        \includegraphics[width=\linewidth]{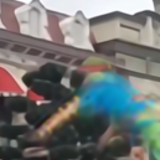} &
        \includegraphics[width=\linewidth]{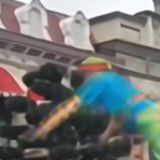} &
        \includegraphics[width=\linewidth]{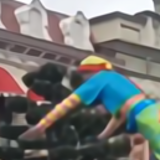}
        \tabularnewline
        
        & 
        \multicolumn{1}{!{\vrule width 1pt}c}{\centering~$V \rightarrow F$} &
        \includegraphics[width=\linewidth]{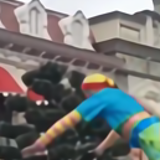} &
        \includegraphics[width=\linewidth]{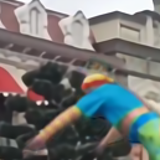} &
        \includegraphics[width=\linewidth]{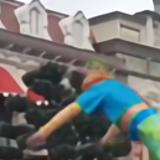} &
        \includegraphics[width=\linewidth]{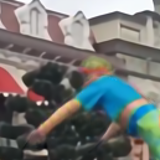} &
        \includegraphics[width=\linewidth]{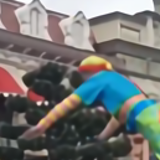}
        \tabularnewline
        
        & 
        \multicolumn{1}{!{\vrule width 1pt}c}{\parbox{0.1\linewidth}{\centering~STVUN (Ours)}} &
        \includegraphics[width=\linewidth]{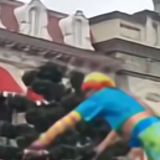} &
        \includegraphics[width=\linewidth]{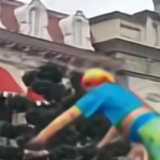} &
        \includegraphics[width=\linewidth]{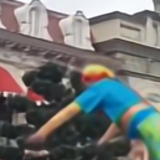} &
        \includegraphics[width=\linewidth]{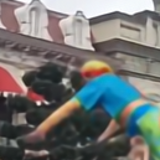} &
        \includegraphics[width=\linewidth]{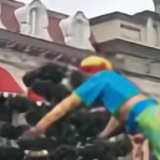}
        \tabularnewline
        
        %% Soccer 
        
        \multirow{3}{*}{\shortstack{\includegraphics[width=\linewidth]{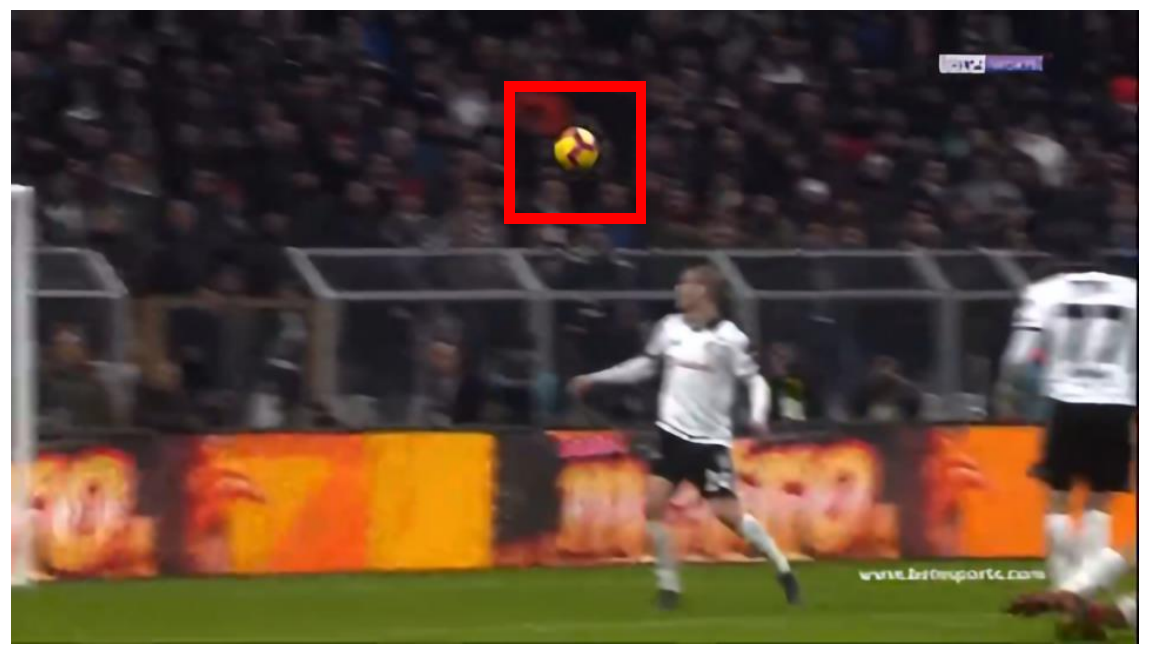}\\Soccer}} & 
        \multicolumn{1}{!{\vrule width 1pt}c}{\centering~$F \rightarrow V$} &
        \includegraphics[width=\linewidth]{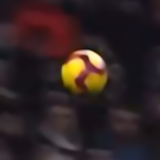} &
        \includegraphics[width=\linewidth]{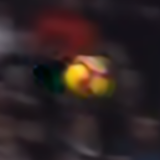} &
        \includegraphics[width=\linewidth]{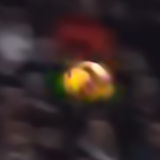} &
        \includegraphics[width=\linewidth]{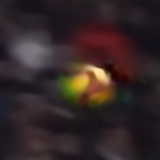} &
        \includegraphics[width=\linewidth]{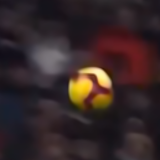}
        \tabularnewline
        
        & 
        \multicolumn{1}{!{\vrule width 1pt}c}{\centering~$V \rightarrow F$} &
        \includegraphics[width=\linewidth]{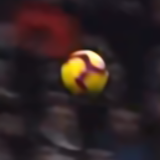} &
        \includegraphics[width=\linewidth]{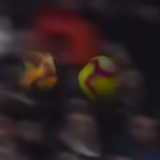} &
        \includegraphics[width=\linewidth]{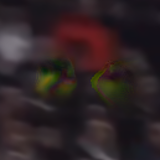} &
        \includegraphics[width=\linewidth]{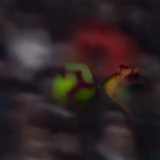} &
        \includegraphics[width=\linewidth]{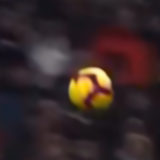}
        \tabularnewline
        
        & 
        \multicolumn{1}{!{\vrule width 1pt}c}{\parbox{0.1\linewidth}{\centering~STVUN (Ours)}} &
        \includegraphics[width=\linewidth]{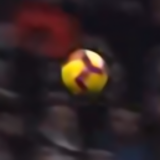} &
        \includegraphics[width=\linewidth]{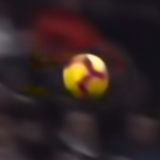} &
        \includegraphics[width=\linewidth]{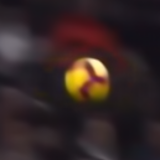} &
        \includegraphics[width=\linewidth]{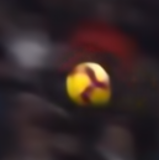} &
        \includegraphics[width=\linewidth]{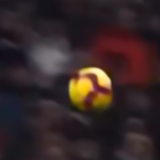}
        \tabularnewline
        
        %% Racing 
        
        \multirow{3}{*}{\shortstack{\includegraphics[width=\linewidth]{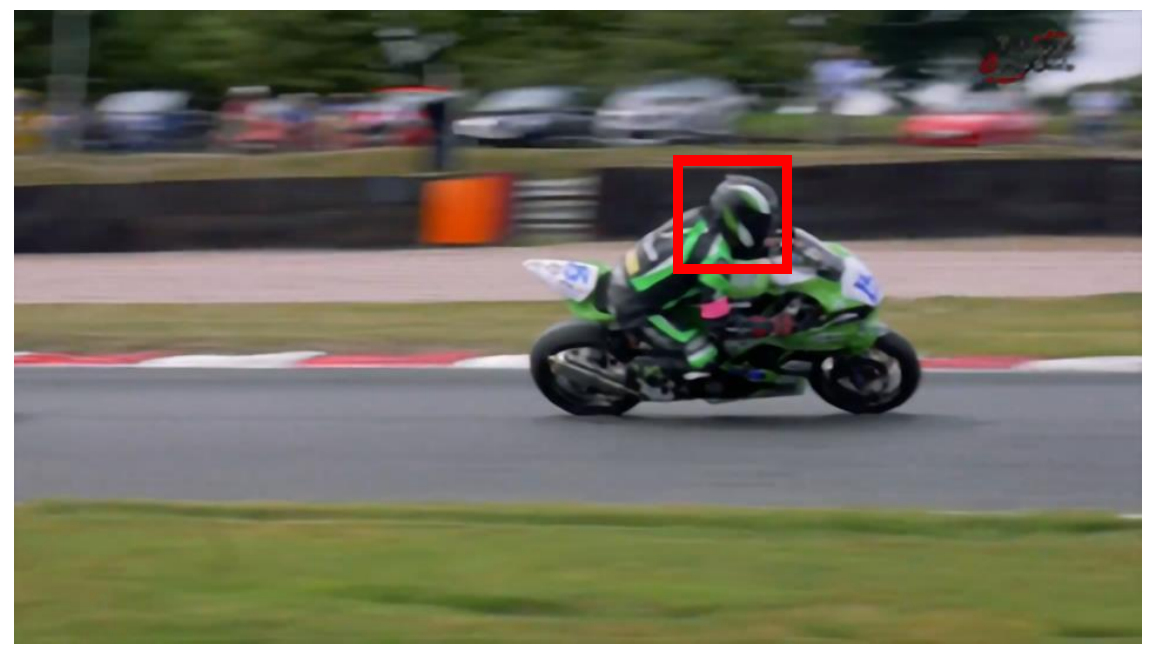}\\Racing}} & 
        \multicolumn{1}{!{\vrule width 1pt}c}{\centering~$F \rightarrow V$} &
        \includegraphics[width=\linewidth]{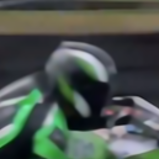} &
        \includegraphics[width=\linewidth]{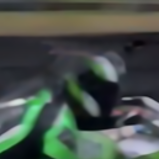} &
        \includegraphics[width=\linewidth]{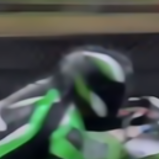} &
        \includegraphics[width=\linewidth]{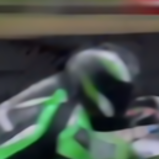} &
        \includegraphics[width=\linewidth]{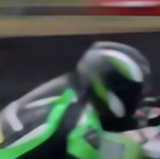}
        \tabularnewline
        
        & 
        \multicolumn{1}{!{\vrule width 1pt}c}{\centering~$V \rightarrow F$} &
        \includegraphics[width=\linewidth]{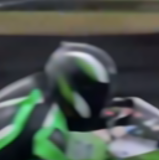} &
        \includegraphics[width=\linewidth]{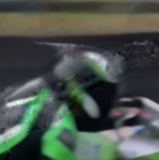} &
        \includegraphics[width=\linewidth]{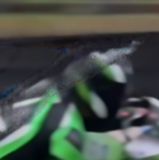} &
        \includegraphics[width=\linewidth]{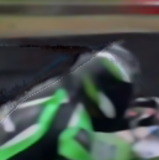} &
        \includegraphics[width=\linewidth]{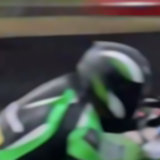}
        \tabularnewline
        
        & 
        \multicolumn{1}{!{\vrule width 1pt}c}{\parbox{0.1\linewidth}{\centering~STVUN (Ours)}} &
        \includegraphics[width=\linewidth]{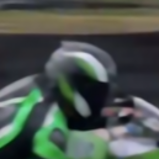} &
        \includegraphics[width=\linewidth]{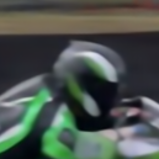} &
        \includegraphics[width=\linewidth]{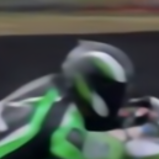} &
        \includegraphics[width=\linewidth]{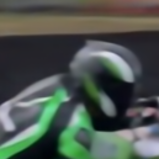} &
        \includegraphics[width=\linewidth]{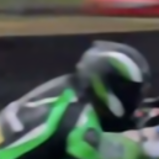}
        \tabularnewline
        
        \end{tabular}
    % \end{tabular}
        \caption{Comparison with baseline methods on STVT dataset.}
        \label{fig:Base_Comparison}
    \end{subfigure}
    
     \vspace{5pt}
    
    \begin{subfigure}[h]{\linewidth}
    \centering
    \begin{tabular}
        {@{}m{0.24\linewidth}@{\hskip2pt}m{0.24\linewidth}@{}m{0.01\linewidth}@{\hskip2pt}m{0.24\linewidth}@{\hskip2pt}m{0.24\linewidth}}
        \centering~FISR &
        \centering~STVUN\\(Ours) &
        &
        \centering~FISR &
        \centering~STVUN\\(Ours) \tabularnewline
        
        \includegraphics[width=\linewidth]{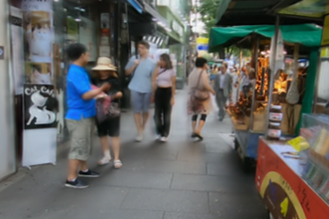} &
        \includegraphics[width=\linewidth]{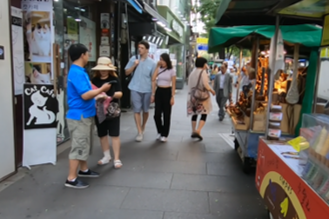} &
        &
        \includegraphics[width=\linewidth]{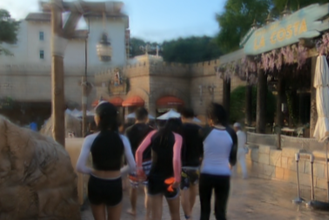} &
        \includegraphics[width=\linewidth]{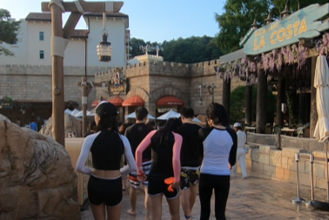} 
    \end{tabular}
        \caption{Comparison with FISR on REDS-VTSR testset. }
        \label{fig:FISR_Comparison}
    \end{subfigure}
    
    \vspace{2pt}
    
    \caption{Visual comparisons of the space-time upsampling results. 
	 In (a), we generate a total of 5 frames that consist of 2 space upsampling and 3 intermediate frames.
	\textit{Parade\_1}, \textit{Soccer}, \textit{Racing} scenes are used in our \textit{STVT} dataset.
	In (b), we generate one intermediate frame.
	\textit{002} and \textit{007} in REDS-VTSR testset are used.
    % In both tasks, our method delivers more visually pleasing results. 
   }

	\label{fig:STVUN_qualitivity}
    
\end{figure}

Additionally, we compare our model with FISR \cite{kim2020fisr} in \Tref{tab:FISR}.
As the upsampling factor of FISR is space $\times2$ and time $\times2$, we train our network with the same settings.
Note that only the number of output channels of the last convolutional layer in the decoder is changed.
% As our environment is different from FISR, we retrain our model for space $\times2$ and time $\times2$.
% As can be seen in \Tref{tab:FISR}, our method outperforms FISR by a large margin in all metrics.
As can be seen in \Tref{tab:FISR}, our method outperforms FISR by a large margin except for REDS-VTSR.
In addition, our method runs faster than FISR with fewer parameters.

\fref{fig:STVUN_qualitivity} visually compares our method with the two baseline methods and FISR.
In \fref{fig:Base_Comparison}, we generate multiple frame ($T_{in}=0.25, 0.5, 0.75$) in-between two input frames.
To better illustrate the results, we enlarge the corresponding red areas.
As STVT dataset has a large motion, two baseline methods have difficulty in handling the large motion.
%and generate awkward results.
In the soccer scene, $F\rightarrow V$ shows more pleasing result than $V\rightarrow F$ because it is easier to estimate the motion at smaller input size. 
% In general, however, most $V\rightarrow F$ produces more pleasing results than $F\rightarrow V$.
Except for the large motion scene, $V \rightarrow F$ is clearer than $F \rightarrow V$.
Overall, our method is more accurate in estimating the motion and shows less artifacts. 
In \fref{fig:FISR_Comparison}, we generate one intermediate frame ($T_{in}=0.5$) for comparison with FISR. 
% The results in FISR show blurry artifact due to wrong motion estimation, but ours restore sharper edge details.
% The results in FISR show ghost artifact due to wrong motion estimation, but on REDS-VTSR, they are close to the ground-truth which makes PSNR more higher than ours.
% The results in FISR show ghost artifact due to wrong motion estimation.
% On the other hand, ours have more sharper details than FISR.
The results in FISR show ghost artifact due to wrong motion estimation, but ours restore sharper edge details.
% However, since ours do not align motion exactly same with the ground-truth, ours has lower PNSR value than FISR.  
% However, due to the unnatural movement of REDS-VTSR testset, the motion in natural movement at $T_{in}=1/2$ is not exactly same with it.
% However, due to the unnatural movement of REDS-VTSR testset, the motion of general case at $T_{in}=1/2$ is not exactly same with ground-truth.
However, due to the unnatural movement of REDS-VTSR testset, most center frames are not in the middle of the front and rear frames.
So, only for this testset, the blurry results of FISR reduce average pixel error than ours.
% 여기서 FISR 이 오히려 ghost effect 때문에 ~~ 수치가 더 높게 된다.
% ours do not align object exactly same with the ground-truth, which makes PNSR value lower than FISR.
% But on REDS-VTSR, they are close to the ground-truth which makes PSNR more higher than ours.
% Although ours do not align motion exactly same with the ground-truth, the results have more sharper details than FISR. 
We recommend watching our demo video in the supplementary material to see the difference more clearly.
 
Beside the STVUN, our network can be used for VSR.
As our main objective is the space-time upsampling, the experiments on VSR will be shown in the supplementary material.

%% Ablation studies
\begin{table}[t]
\centering
\caption{Ablation studies on the EFST and our network structure.
STVT dataset is used for comparison.}
\vspace{1pt}
	\begin{tabular}
		{P{0.18\linewidth}||P{0.18\linewidth}@{}P{0.18\linewidth}@{}P{0.18\linewidth}@{}P{0.19\linewidth}}
		\hline
		& w/o EFST & w/o $D$ & w/ A\&F & Ours\\
		\hline
		\hline
% 		\multirow{2}{*}{\vspace{-2pt}\shortstack[l]}
		PSNR/SSIM & 27.06/0.8613 & 27.15/0.8615 & 27.20/\textbf{0.8652} & \textbf{27.23}/0.8644 \\
		\#Params  & \textbf{30.5M} & 30.9M & 32.4M & 30.9M \\
		Running Time & \textbf{0.27s} & 0.30s & 0.75s & 0.30s \\
		\hline
	\end{tabular}
\label{tab:Ablation}
    \vspace{-10pt}
\end{table}

\subsection{Ablation Studies}
We conduct ablation studies to investigate the contribution of EFST and our network design.
\Tref{tab:Ablation} summarizes the ablation results.
First, we test our model without EFST (w/o EFST), which means only the early fusion is used to fuse input features.
This test demonstrates the effectiveness of EFST as it shows that our final model improves the performance without the large difference in running time.

% runtime is reduced by 0.03s but PSNR also decreased by 0.17dB.
% Instead of using the decoded features, EFST features is directly forwarded to space-time upsampling task in Model B.
To show the effectiveness of using decoded features $D^i$ for space-time upsampling, we test our model when EFST features is used instead (w/o $D$).
The performance gain shows learned features from space upsampling enhance the space-time upsampling results, indicating the decoder learn more rich information from EFST features.

We also evaluate our model with the explicit alignment (w/ A\&F).
The overall structure is the same as our proposed method except for the alignment and fusion parts.
% As our EFST is for feature fusion without explicit alignment, we use two modules in EDVR~\cite{wang2018non}.
We use two modules in EDVR~\cite{wang2018non} -- Pyramid, Cascading and Deformable Convolution (PCD) for the alignment and TSA for the feature fusion.
As the explicit alignment process is a computational burden, it increases the running time by about 2.5 times.
But the performance gap is minor, demonstrating EFST can effectively fuse features without explicit motion compensation..

\begin{figure}[t]
% 	\vspace{5pt}
	%\centering
	\begin{center}
		\includegraphics[width=0.98\linewidth]{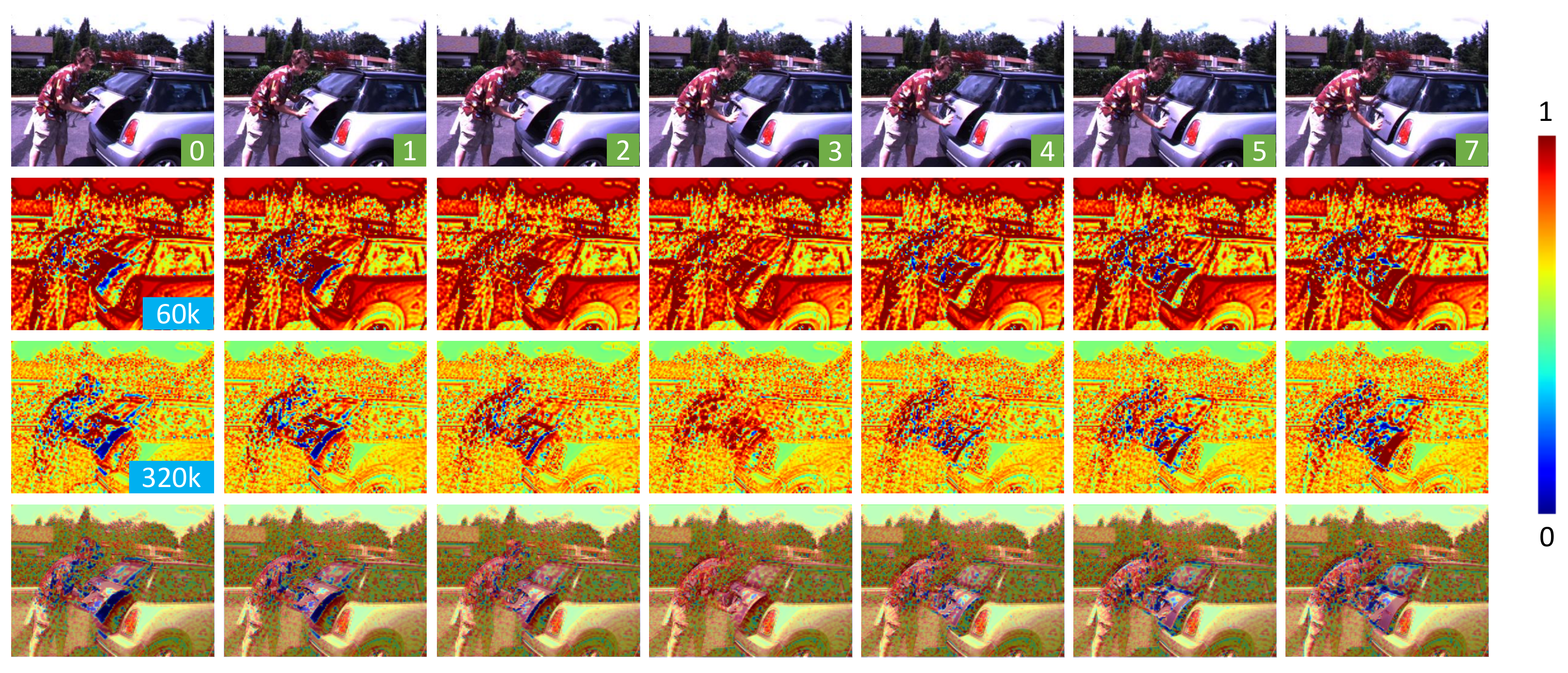}
	\end{center}
	\vspace{-15pt}
	\caption{We visualize the confidence score $s^i$.
	Green box numbers are the time index, and the blue box numbers are the number of iterations.
	High confidence score is shown in red, and this means to be more referred and dark blue is vice versa.
	In the last row, we overlap results of the input and color map of 320k iterations.
	Zoom in to see better visualization.
	}
	\label{fig:Colormap}
	\vspace{-10pt}
\end{figure}

\fref{fig:Colormap} shows the visualization of the confidence score to analyze how confidence score changes with learning in EFST.
% In the early iteration, since $E^i$ is less learned, features from all frames are needed (in dark red).
In the early stages of training, confidence scores are ambiguous to determine where to concentrate more. 
Therefore, the overall scores are high and shown in dark red. 
As the learning progresses, the confidence score gets the ability to determine the important parts among all inputs.
High confidence scores remain for the regions which are helpful for reconstructing the center frame.
On the other hand, occluded regions such as under the trunk lid in frame 0 have low confidence score because they are unnecessary for reconstructing center frame.
It demonstrates that our confidence score effectively fuses features from all frames without explicit alignment.

\section{Conclusion}
In this paper, we present a deep Space-Time Video Upsampling Networks (STVUN) for joint space-time video upsampling by merging VSR and FI network efficiently.
This task has many practical applications, yet a challenging task as the network has to perform two tasks in an efficient manner.
In addition, we propose Early Fusion with Spatio-Temporal weights (EFST) modules that learns to fuse information by considering spatio-temporal relationship without any explicit alignment.
Our network can generate visually pleasing results with reduced computational time ($\times 7$) and number of parameters (30\%) compared to sequentially connected VSR and FI networks.
Our method also outperforms a previous space-time upsampling task by a large margin.

\newpage
\bibliographystyle{splncs04}
\bibliography{egbib}
\end{document}